\documentclass[letterpaper, 10 pt, conference]{ieeeconf}  

\IEEEoverridecommandlockouts                              

\usepackage{amsmath}
\usepackage{amssymb}
\usepackage{amsfonts}
\usepackage{latexsym}
\usepackage{lipsum}
\usepackage{multicol}
\usepackage{accents}
\usepackage{harpoon}
\usepackage{wrapfig}
\usepackage{tabularx}
\usepackage{euscript}
\usepackage{color}
\usepackage{xcolor}
\usepackage{graphicx}
\usepackage{float}
\usepackage[caption = false]{subfig}	
\usepackage{epsfig}
\usepackage{hyperref}
\usepackage{setspace}
\usepackage{fancyhdr}
\usepackage{textcomp}
\usepackage{cite}
\usepackage{epstopdf}
\usepackage{subfig}
\usepackage{verbatim}
\usepackage[left=0.75in,top=1in,right=0.75in,bottom=0.75in]{geometry}
\DeclareGraphicsExtensions{.pdf,.eps,.jpg}
\graphicspath{{Figures/}}

\usepackage{pifont}
\usepackage{latexsym}
\usepackage{psfrag}
\usepackage{stmaryrd}

\newcounter{mynum2}

\title{\LARGE \bf
A Distributed Control Framework for a Team of Unmanned Aerial Vehicles for Dynamic Wildfire Tracking
}

\author{Huy X. Pham, Hung M. La, David Feil-Seifer, and Matthew Deans
\thanks{*This work is supported by Nevada NASA Research Infrastructure Development Program under the 2016 Seed Grant Award.}
\thanks{Huy Pham is a PhD student, and Dr. Hung La is the director of the Advanced Robotics and Automation
(ARA) Laboratory. Dr. David Feil-Seifer is an Assistant Professor at Department of Computer Science and Engineering, University
of Nevada, Reno, NV 89557, USA. Dr. Matthew Deans is with NASA Ames Research Center, Moffett Field, CA 94035. Corresponding author: Hung La, email: {\tt\small hla@unr.edu}}
}

\begin{document}

\maketitle
\thispagestyle{empty}
\pagestyle{empty}

\begin{abstract}
Wildland fire fighting is a very dangerous job, and the lack of information of the fire front is one of main reasons that causes many accidents. Using unmanned aerial vehicle (UAV) to cover wildfire is promising because it can replace human in hazardous fire tracking and save operation costs significantly. In this paper we propose a distributed control framework designed for a team of UAVs that can closely monitor a wildfire in open space, and precisely track its development. The UAV team, designed for flexible deployment, can effectively avoid in-flight collision as well as cooperate well with other neighbors. Experimental results are conducted to demonstrate the capabilites of the UAV team in covering a spreading wildfire.
\end{abstract}

\section{Introduction}\label{S.intro}
\begin{figure}[t]
\centering
\includegraphics[width=0.8\columnwidth]{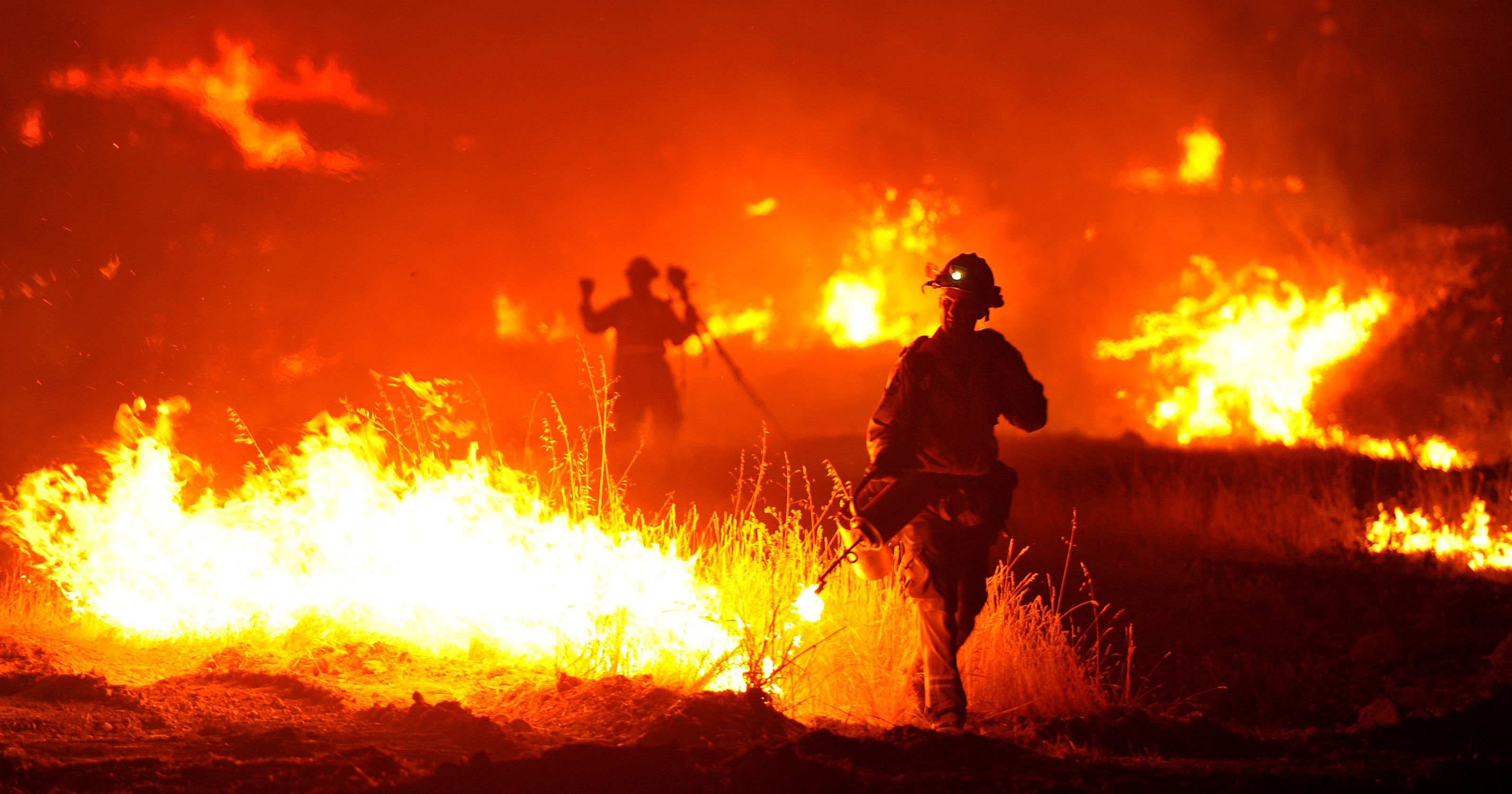}
  \caption{A wildfire outbreaks in California. Firefighting is really dangerous without continuous fire fronts growth information. Courtesy of USA Today.}
  \label{F.Fightfighters}
\vspace{-15pt}
\end{figure}

Wildfire is well-known for their destructive ability to inflict massive damages and disruptions. According  to the U.S. Wildland Fire, an average of 70000 wildfires annually burn around 7 million acres of land and destroy more than 2600 structure~\cite{nifc2017}. Wildfire fighting is usually dangerous and time sensitive. The lack of information about the current state and the dynamic evolution of fire contributes to many accidents~\cite{martinez2008computer}. Firefighters may easily lose their life if the fire unexpectedly propagates over them (figure \ref{F.Fightfighters}). Therefore, there is an urgent need to locate the wildfire correctly~\cite{stipanivcev2010advanced}, and it is even more important to precisely cover the development of the fire and track its spreading boundaries~\cite{SujitP_2007}. The more information regarding the fire spreading areas collected, the better the strategies we can formulate to evacuate people and properties out of the danger zones, as well as effectively prevent the fire from escalating to other sections.

Using Unmanned Aircraft Systems (UAS), also called Unmanned Aerial Vehicles (UAV) or drones, to assist wildfire fighting and other natural disaster relief is very promising. They can assist human in hazardous fire tracking tasks and replace the use of manned helicopters, while saving sizable operation costs in comparison with traditional methods~\cite{MerinoL_2006}~\cite{cruz2016efficient}. However, research that discusses the application of UAVs in assisting fire fighting remains limited~\cite{yuan2015survey}.

Although current UAV technology has not fully matured, recent advancement allows UAVs to host a wide range of sensing capabilities. Accurate UAV-based fire detection has been thoroughly demonstrated in current research. Merino et al.~\cite{MerinoL_2006} proposed a cooperative perception system featuring infrared, visual camera, and fire detectors mounted on different UAV types. The system can precisely detect and estimate fires location. Yuan et al.~\cite{YuanC_2015} developed a fire detection technique by analyzing fire segmentation in different color spaces. An efficient algorithm was proposed in~\cite{cruz2016efficient} to work on UAV with low-cost cameras, using color index to distinguish fire from smoke, steam and forest environment under fire, even in early stage. Merino et al.~\cite{merino2012unmanned} utilized a team of UAVs to collaborate together to obtain fire front shape and position. In these works, camera plays a crucial role in capturing the raw information for higher level detection algorithms.

\begin{figure}[b]
\centering
\includegraphics[width=0.8\columnwidth]{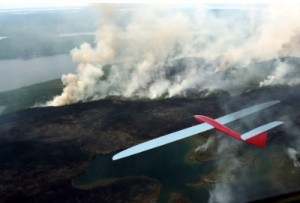}
  \caption{A UAV monitoring a wildfire. Courtesy of NASA.}
  \label{F.UAVs}
\vspace{-15pt}
\end{figure}

\begin{figure*}[!t]
\normalsize
\setcounter{mynum2}{\value{equation}}
\begin{equation}
\label{2.truefarsite}
\begin{aligned}
	X_{t} &= \frac{a^{2}\cos\Theta(x_{s}\sin\Theta + y_{s}\cos\Theta) - b^{2}\sin\Theta(x_{s}\cos\Theta - y_{s}\sin\Theta)}{\sqrt{b^{2}(x_{s}\cos\Theta + y_{s}\sin\Theta) - a^{2}(x_{s}\sin\Theta - y_{s}\cos\Theta}}  
		+ c\sin\Theta \\
	Y_{t} &= \frac{-a^{2}\sin\Theta(x_{s}\sin\Theta + y_{s}\cos\Theta) - b^{2}\cos\Theta(x_{s}\cos\Theta - y_{s}\sin\Theta)}{\sqrt{b^{2}(x_{s}\cos\Theta + y_{s}\sin\Theta) - a^{2}(x_{s}\sin\Theta - y_{s}\cos\Theta}}  
		+ c\cos\Theta,\\
\end{aligned}
\end{equation}
\setcounter{equation}{1}
\hrulefill
\vspace{-15pt}
\end{figure*}

Research groups also underline the importance of coordination between the UAVs to have a better coverage of the fire, as it will allow more information collected and larger areas covered. Maza et al.~\cite{MazaI_2011} provided an distributed decisional architecture framework for multi-UAV applications in disaster management. In~\cite{CasbeerD_2005}, a multiple UAVs are commanded to track a spreading fire using checkpoints calculated based on visual images of the fire perimeter. In addition, another research group~\cite{kumar2011cooperative} proposed algorithms using artificial potential field to control a team of UAVs in two separated tasks: track the boundary of a wildfire and suppress it. A centralized optimal task allocation problem is formulated in~\cite{phan2008cooperative} to generate a set of waypoints for UAVs for shortest path planning. 

However, to the best of the authors' knowledge, most of the above mentioned work do not cover the behaviors of their system when the fire is spreading. Works in~\cite{CasbeerD_2005} and~\cite{phan2008cooperative} centralized the decision making, thus potentially overloaded in computation and communication when the fire in large scale demands more UAVs. The team of UAVs in~\cite{kumar2011cooperative} can continuously track the boundary of the spreading fire but largely depends on the accuracy of the modeled shape function of the fire in control design. In this paper, we propose a decentralized control algorithm for a team of UAVs that can autonomously and actively track the fire spreading boundaries in a distributed manner, without dependency on the wildfire modeling. The UAVs can effectively share the vision of the field, while maintaining safe distance in order to avoid in-flight collision. Moreover, during tracking, the proposed algorithm can allow the UAVs to increase image resolution captured on the border of the wildfire.

The rest of the paper is organized as follows: section 2 discusses about how we model the wildfire spreading as an objective for this paper. In section 3, the wildfire tracking problem is formulated with clear objectives. In section 4, we propose a control design capable of solving the problem. A simulation scenario on Matlab are provided in section 5. Finally, we draw a conclusion, and suggest directions for future work.

\begin{figure*}[t]
 \centering
	\subfloat[t = 0]{\includegraphics[width=0.25\textwidth]{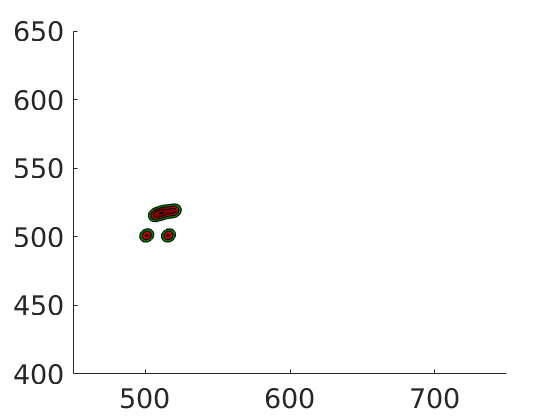}}
	\subfloat[t = 1000]{\includegraphics[width=0.25\textwidth]{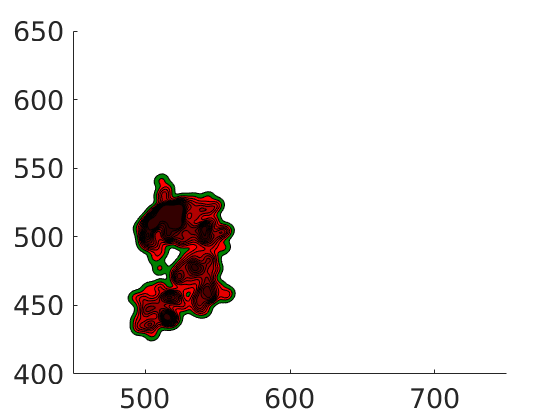}}
	\subfloat[t= 3000]{\includegraphics[width=0.25\textwidth]{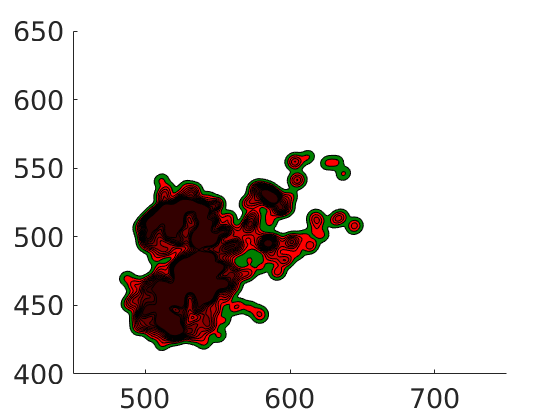}}
	\subfloat[t = 6000]{\includegraphics[width=0.25\textwidth]{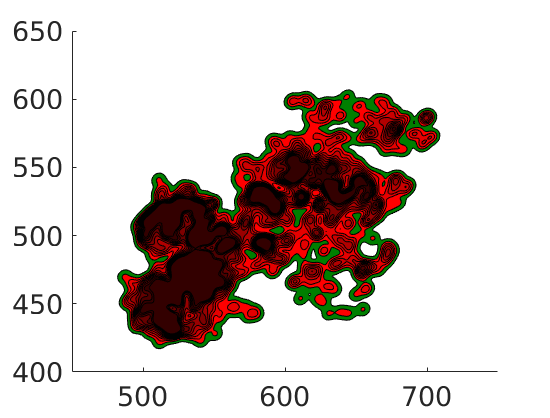}}
  \caption{Simulation result shows a wildfire spreading at different time steps.}
  \label{F.wildfire}
\vspace{-15pt}
\end{figure*}

\section{Wildfire Modeling}\label{S.wildfire}

Wildfire simulation has attracted significant research efforts over the past decades, due to the potential in predicting wildfire spreading. The core model of existing fire simulation systems is the fire spreading propagation~\cite{glasa2011note}. Rothermel in 1972~\cite{rothermel1972mathematical} developed basic fire spread equations to mathematically and empirically calculate rate of speed and intensity. Richards~\cite{richards1990elliptical} introduced a technique to estimate fire fronts growth using an elliptical model. These previous research were later developed further by Finney~\cite{finney2004farsite} and became a well-known fire growth model called Fire Area Simulator (FARSITE). Among existing systems, FARSITE is the most reliable model~\cite{williams2014modeling}, and widely used by federal land management agencies such as USDA Forest Service. However, in order to implement the model precisely, we need significant information regarding geography, topography, conditions of terrain, fuels, and weather. To focus on the scope of multi-UAV control rather than pursuing an accurate fire growth model, in this paper we modify the fire spreading propagation in FARSITE model to describe the fire fronts growth in a simplified model. We also make the following assumptions:

\begin{itemize}
\item the model will be implemented for a discrete grid-based environment;
\item the steady-state rate of spreading is already calculated for each grid;
\item only the fire front points spread.
\end{itemize}

Originally, the equation for calculating the differentials of spreading fire front proposed in~\cite{richards1990elliptical} and~\cite{finney2004farsite} as equation (\ref{2.truefarsite}), where $X_{t}$ and $Y_{t}$ are the differentials, $\Theta$ is the azimuth angle of the wind direction and y-axis ($0\leq \Theta \leq 2\pi$). $\Theta$ increases following clock-wise direction. $a$ and $b$ are the length of semi-minor and semi-major axes of the elliptical fire shape growing from one fire front point, respectively. $c$ is the distance from the fire source (ignition point) to the center of the ellipse. $x_{s}$ and $y_{s}$ are the orientation of the fire vertex. We simplify the equation (\ref{2.truefarsite}) to only retain the center of the new developed fire front as follows:
\begin{equation}\label{2.firespread}
\begin{aligned}
	X_{t} &= c\sin\Theta \\
	Y_{t} &= c\cos\Theta. \\ 
\end{aligned}
\end{equation}
We use equation from Finney~\cite{finney2004farsite} to calculate $c$ according to the set of equations (\ref{2.firespread}) as follows:
\begin{equation}\label{E.rel_distance}
\begin{aligned}
	\ LB &= 0.936 e^{0.2566U} + 0.461 e^{- 0.1548U}  - 0.397 \\
	\ HB &= \frac{LB + (LB^{2} - 1)^{0.5}}{LB - (LB^{2} - 1)^{0.5} }\\	
	\ c &= \frac{R - \frac{R}{HB}}{2},\\
\end{aligned}
\end{equation}
where $R$ is the steady-state rate of fire spreading. $U$ is the scalar value of mid-flame wind speed, which is the wind speed at the ground. It can be calculated from actual wind speed value after taking account of the wind resistance by the forest. The new fire front location after time step $\delta t$ is calculated as:
\begin{equation}\label{2_location}
\begin{aligned}
	x_{f}(t+1) &= x_{f}(t) + \delta tX_{t}(t)\\
	y_{f}(t+1) &= y_{f}(t) + \delta tY_{t}(t).\\
\end{aligned}
\end{equation}
Additionally, in order to simulate the intensity caused by fire around each fire front source, we also assume that each fire front source would radiate energy to the surrounding environment resembling a multivariate normal distribution probability density function of its coordinates $x$ and $y$. Assuming linearity, the intensity of each point in the field is a linear summation of intensity functions caused by multiple fire front sources. Moreover, due to the exhaustion of the fuel, the intensity is subject to a decay rate $\lambda$ over the time. Therefore, we have the following equation describing the intensity of each point in the wildfire caused by a number of $k$ sources:
\begin{equation}\label{2.multivariate}
\begin{aligned}
	\ I(x, y) &= \sum^{k}_{i = 1}\frac{1}{2\pi\sigma_{x_{i}}\sigma_{y_{i}}}e^{-\frac{1}{2}[\frac{(x-\mu_{x_{i}})^2}{\sigma_{x_{i}}^2}+\frac{(y-\mu_{y_{i}})^2}{\sigma_{y_{i}}^2}]}e^{-\lambda t}, \\
\end{aligned}
\end{equation}
where $I(x, y)$ is the intensity of the fire at point $(x, y)$, $(\mu_{x_{i}}, \mu_{y_{i}})$ coincide with the location of the heat source $i$, and $(\sigma_{x_{i}}, \sigma_{y_{i}})$ are deviations. The point closer to the heat source has a higher level of intensity of the fire. Figure \ref{F.wildfire} represents the simulated wildfire spreading from original source (a) until $t = 6000$ time steps (d). The simulation assumes the wind flows north-east with direction is normally distributed ($\mu_{\Theta} = \frac{\pi}{8}, \sigma_{\Theta} = 1$), midflame adjusted wind speed is also normally distributed ($\mu_{U} = 5, \sigma_{u} = 2$). The decay rate is $\lambda = 0.01$. The green area depicts the boundary with forest field, while red area represents the fire. The brighter red color area illustrates the outer of the fire and regions near the boundary where the intensity is lower. The darker red colors show the area in fire with high intensity.

It should be noted that in this paper, the fire growth model solely serves to demonstrate a continuous changing environment and simulates an objective for the team of UAVs to cover. The accuracy of the model will not affect the performance of our distributed control algorithm.

\section{Problem formulation}\label{S.formulation}

In this section, we translate our motivation into a formal problem formulation. Our objective is to control a team of multiple UAVs for collaboratively covering a wildfire and tracking the fire front propagation. By covering, we mean to let the UAVs take multiple pictures of the affected location. We assume that the fire happens in a known section of a forest, where the priori information regarding the location of any specific point are made available. Suppose that when a wildfire happens, its estimated location is notified to the UAVs. A command is then sent to the UAV team allowing them to start. The team needs to satisfy the following objectives:

\begin{itemize}
\item \textit{Deployment objective}: The UAVs can take flight from the deployment depots to the initially estimated wildfire location. 
\item \textit{Coverage and tracking objective}: Upon reaching the reported fire location, the team will spread out to cover the entire wildfire from a certain altitude. The UAVs then follow and track the development of the fire fronts. When following the expanding fire fronts of the wildfire, some of the UAV team may lower their altitude to increase the image resolution of the fire boundary, while the whole team tries to maintain a complete view of the wildfire.
\item \textit{Collision avoidance objective}: Because the number of UAVs can be large (i.e. for sufficient coverage a large wildfire), it is important to ensure that the participating UAVs are able to avoid in-flight collisions with other UAVs.
\end{itemize}

\begin{figure}[h]
 \centering
  \includegraphics[width=0.8\columnwidth]{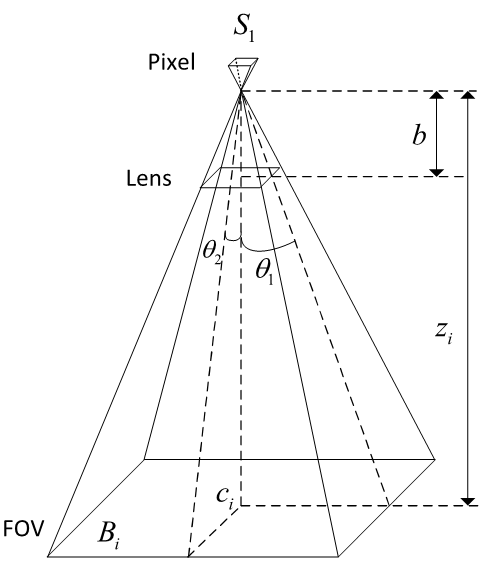}
  \caption{Rectangular field of view of a UAV, with half-angles $\theta_{1}$, $\theta_{2}$. Each UAV will capture the area under its field of view using its camera, and record the information into a number of pixels.}
  \label{F.fieldofview}
\vspace{-15pt}
\end{figure}

Assume that each UAV equipped with localization devices (such as GPS and IMU), and identical downward-facing cameras capable of detecting fire. Each camera has a rectangular \textit{field of view} (FOV). When covering, the camera and its FOV form a pyramid with half-angles $\theta^{T} =[\theta_{1}, \theta_{2}]^{T}$ (see figure \ref{F.fieldofview}). Each UAV will capture the area under its FOV using its camera, and record the information into a number of pixels. We also assume that a UAV can communicate and exchange information with another UAV if it remains inside a communication sphere with radius $r$ (see figure \ref{F.potentialfieldneighbors}). 

We define the following variables that will be used throughout this paper.
Let $N$ denote the set of the UAVs. Let $p_{i} = [c_{i}^{T}, z_{i}]^{T}$denote the pose of a UAV $i \in N$. In which, $c_{i}^{T} = [x_{i}, y_{i}]^{T}$ indicates the lateral coordination, and $z_{i}$ indicates the altitude. Let $B_{i}$ denote the set of points that lie inside the field of view of UAV $i$. Let $l_{k}, k = 1:4$ denotes each edge of the rectangular FOV. Let $n_{k}, k = 1:4$ denote the outward-facing normal vectors of each edge, where $n_{1} = [1, 0]^{T}$,  $n_{2} = [0, 1]^{T}$,  $n_{3} = [-1, 0]^{T}$,  $n_{4} = [0, -1]^{T}$. We then define the objective function for each task of the UAV team.

\subsection{Deployment objective}
The UAVs can be deployed from depots distributed around the forest, or from a forest firefighting department center. Upon receiving the report of a wildfire happening, the UAVs are commanded to start and move to the point where the location of the fire initially estimated. We call this point a rendezvous point $p_{r} = [p_{x}, p_{y}, p_{z}]^{T}$. The UAVs would keep moving toward this point until they can detect the wildfire inside their FOV.

\noindent

\subsection{Coverage and tracking objective}
Let $Q(t)$ denote the wildfire varying over time $t$ on a plane. From the relationship between object and image distance through a converging lens in classic optics, we can easily calculate the FOV area that a UAV covers (see figure \ref{F.fieldofview}) as follows:

\begin{equation}\label{3_areaperpixel}
\begin{aligned}
	\ f(p_{i},q) = \frac{S_{1}}{b^{2}}(b-z_{i})^{2}, \forall q \in B_{i}, \\
\end{aligned}
\end{equation}
\noindent

where $q^{T} = [q_{x}, q_{y}]^{T}$ is the coordination of a given point that belongs to $Q(t)$, $S_{1}$ is the area of one pixel of a camera, and $b$ denotes the focal length. Since each camera has limited number of pixels to capture an image, it will provide one snapshot of the wildfire with lower resolution when covering it in a bigger FOV. We desire to provide higher-resolution images of the fire border by minimizing the information captured by the pixels. Schwager et al.~\cite{schwager2011eyes} formulated an objective function to minimize the information over total number of pixels from $n$ cameras to cover a static field $Q$, with respect to its strategic level of importance, denoted by importance function $\phi(q)$ as:

\begin{equation}\label{3_Schwagersobjective}
\begin{aligned}
	\min H(p_{1},...,p_{n}) &= \int_{Q}(\sum_{i\in N_{q}}f(p_{i},q)^{-1}+w^{-1})^{-1}\phi(q)dq,
\end{aligned}
\end{equation}

\noindent
where $w$ is a constant to represent some priori knowledge of the environment, $N_{q}$ is the set of UAVs that include the point $q$ in their FOVs. For a point $q$ to lie on or inside the FOV of a UAV $i$, it must satisfy the following condition:

\begin{equation}\label{3_coveragecondition}
\begin{aligned}
	\frac{||q - c_{i}||}{z_{i}} \leq tan \theta. \\
\end{aligned}
\end{equation}
When two UAVs have one or more points in common, they will become \textit{coverage neighbors}. 

We will adapt the objective function (\ref{3_Schwagersobjective}) so that the UAVs will try to cover the field in the way that considers the region around the border of the fire more important. First, we consider that each fire front radiates a heat aura, as described in equation (\ref{2.multivariate}), section \ref{S.wildfire}. Obviously, the border region of each fire front has the least heat energy, while the center of the fire front has the most intense level. We assume that the UAVs equipped with infrared camera allowing them to sense different color spectra with respect to the levels of fire heat intensity. Furthermore, the UAVs are assumed to have installed an on-board fire detection program to quantify the differences in color into varying levels of fire heat intensity~\cite{cruz2016efficient}. Let $I$ denote the varying levels of fire heat intensity, and suppose that the cameras have the same detection range $[I_{min}, I_{max}]$. The desired objective function that weights the fire border region higher than at the center of the fire allows us to characterize the importance function as follows:
\begin{equation}\label{3_phi}
\begin{aligned}
	\phi(q) &= \kappa (I_{max} - I) &= \kappa \Delta I.\\
\end{aligned}
\end{equation}
Note that some regions at the center of the wildfire may have $I = I_{max}$ now become not important. This makes sense because these regions likely burn out quickly, and they are not the goals for the UAV to track. We have the following objective function for wildfire coverage and tracking objective:
\begin{equation}\label{3_coverobjective}
\begin{aligned}
	\min O &= \int_{Q(t)}(\sum_{i\in N_{q}}f(p_{i},q)^{-1}+w^{-1})^{-1}\kappa \Delta Idq.\\
\end{aligned}
\end{equation}

\begin{figure}[b]
\centering
  \includegraphics[width=0.8\columnwidth]{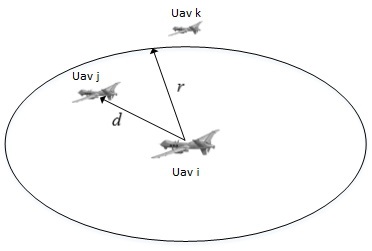}
  \caption{UAV $i$ only communicates with a nearby UAV that flies inside its communication range $r$ (UAV $j$). Each UAV would try to maintain a designed safe distance $d$ to other UAVs in the team.}
  \label{F.potentialfieldneighbors}
\vspace{-15pt}
\end{figure}
\noindent

\subsection{Collision avoidance objective}
The team of UAVs must be able to avoid in-flight collision. In order to do that, a UAV needs to identify its neighbors first. UAV $i$ only communicates with a nearby UAV $j$ that remains inside its communication range (Figure \ref{F.potentialfieldneighbors}), and satisfies the following equation:
\begin{equation}\label{3_flockneighbors}
\begin{aligned}
	\ ||p_{j} - p_{i}|| \leq r, \\
\end{aligned}
\end{equation}
where $r$ is the communication range radius. If equation (\ref{3_flockneighbors}) is satisfied, the two UAVs become \textit{physical neighbors}. For UAV $i$ to avoid other neighbor UAV $j$, they must keep their distance not less than a designed distance $d$:

\begin{equation}\label{3_avoidcollision}
\begin{aligned}
	\ ||p_{j} - p_{i}|| \geq d. \\
\end{aligned}
\end{equation}

\begin{figure}[h]
 \centering
  \includegraphics[width=0.8\columnwidth]{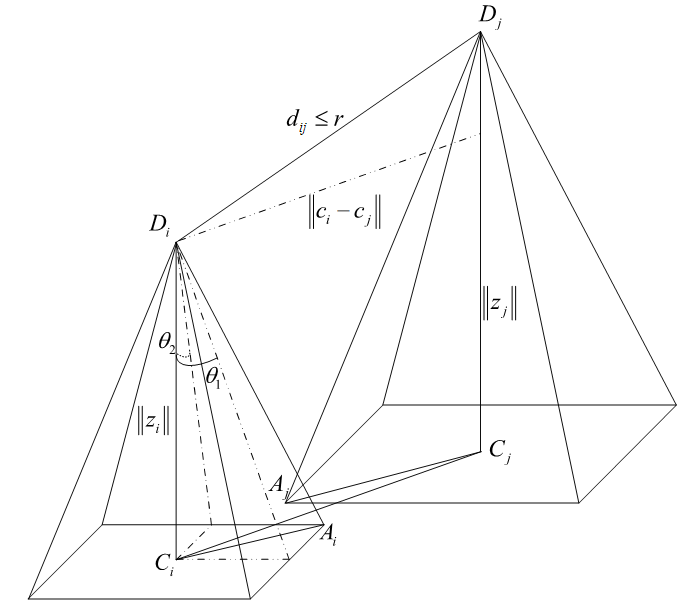}
  \caption{The neighbor relationship between UAV $i$ with pose $[c_{i}^{T}, z_{i}]$ and UAV $j$ with pose $[c_{j}^{T}, z_{j}]$. $d_{ij}$ denotes the distance between them.}
  \label{F.Neighbors}
\vspace{-15pt}
\end{figure}

\subsection{Identify neighbors}
This section provides the link between two definitions of neighborhood. Figure \ref{F.Neighbors} depicts two UAVs in close positions. In order to become physical neighbors, and thus can exchange signals regarding position, the UAV $Dj$ must fall into the communication range of the UAV $Di$, as indicated by (\ref{3_flockneighbors}). We also have:
\begin{equation}\label{3_range}
\begin{aligned}
	\ ||c_{j} - c_{i}||  \leq ||p_{j} - p_{i}||.\\
\end{aligned}
\end{equation}

In order to become sensing neighbors, the UAVs must first become physical neighbors. It also must satisfy a necessary condition:
\begin{equation}\label{3_CAcc}
\begin{aligned}
	\ ||C_{i}A_{i}||  +  ||C_{j}A{j}||  \leq ||c_{i} - c_{j}||,\\
\end{aligned}
\end{equation}

\noindent
where $||C_{i}A_{i}||$ and $||C_{j}A_{j}||$ are computed as:
\begin{equation}\label{3_CAz}
\begin{aligned}
	\ ||C_{i}A_{i}|| &= z_{i}\sqrt{tan^{2}\theta_{1}+tan^{2}\theta_{2}} \\
	\ ||C_{j}A_{j}|| &= z_{j}\sqrt{tan^{2}\theta_{1}+tan^{2}\theta_{2}}, \\
\end{aligned}
\nonumber
\end{equation}
where $\theta_{1}, \theta_{2}$ are constant, as they are half-angles of a FOV which are the same for all the UAVs. From (\ref{3_flockneighbors}), (\ref{3_range}), (\ref{3_CAcc}), and (\ref{3_CAz}), if two UAVs are sensing neighbors, they must satisfy the following necessary condition:
\begin{equation}\label{3_neighborscondition}
\begin{aligned}
	\ z_{i} + z_{j}  \leq \frac{r}{\sqrt{tan^{2}\theta_{1}+tan^{2}\theta_{2}}}.\\
\end{aligned}
\end{equation}

We can use this condition to select the range radius $r$ large enough to guarantee communication among the UAVs that have overlapping field of views. But we must also limit $r$ so that communication overload does not occur as a result of having too many neighbors. Note that (\ref{3_neighborscondition}) only indicates a necessary condition for two UAVs to become sensing neighbors. Even if satisfying that condition, due to the rectangular geometry of the FOV, two UAVs may not become sensing neighbors. In the next section, we propose a unified controller to satisfy all the objectives described in this sections.
\section{Controller Design}\label{S.control}

\begin{figure}[t]
 \centering
  \includegraphics[width=0.8\columnwidth]{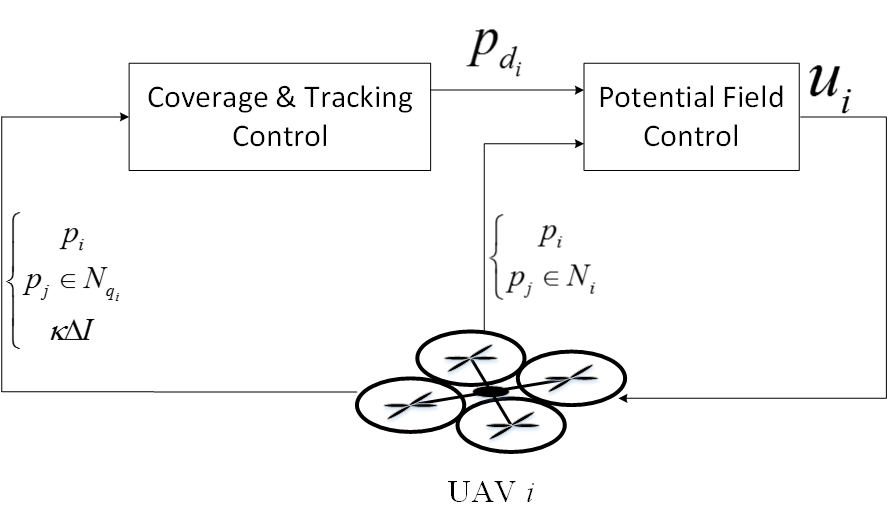}
  \caption{Controller architecture.}
  \label{F.Controller}
\vspace{-15pt}
\end{figure}

Figure \ref{F.Controller} shows our controller architecture for each UAV. Our controller consists of two components. The coverage and tracking component in upper level controls the position of the UAV for wildfire coverage and tracking. The potential field component in lower level controls the UAV to move to desired positions, and to avoid collision with other UAVs by using potential field method.  Upon reaching the wildfire region, the coverage and tracking control component will update the desired position of the UAV to the potential field control component. Assume the dynamics of each UAV is:

\begin{equation}\label{4.dynamics}
\begin{aligned}
	\ u_{i} &= \dot{p_{i}}, \\
\end{aligned}
\end{equation}
\noindent

we can then develop the control equation for each component in the upcoming subsections.

\begin{figure*}[t]
 \centering
	\subfloat[t = 1000]{\includegraphics[width=0.25\textwidth]{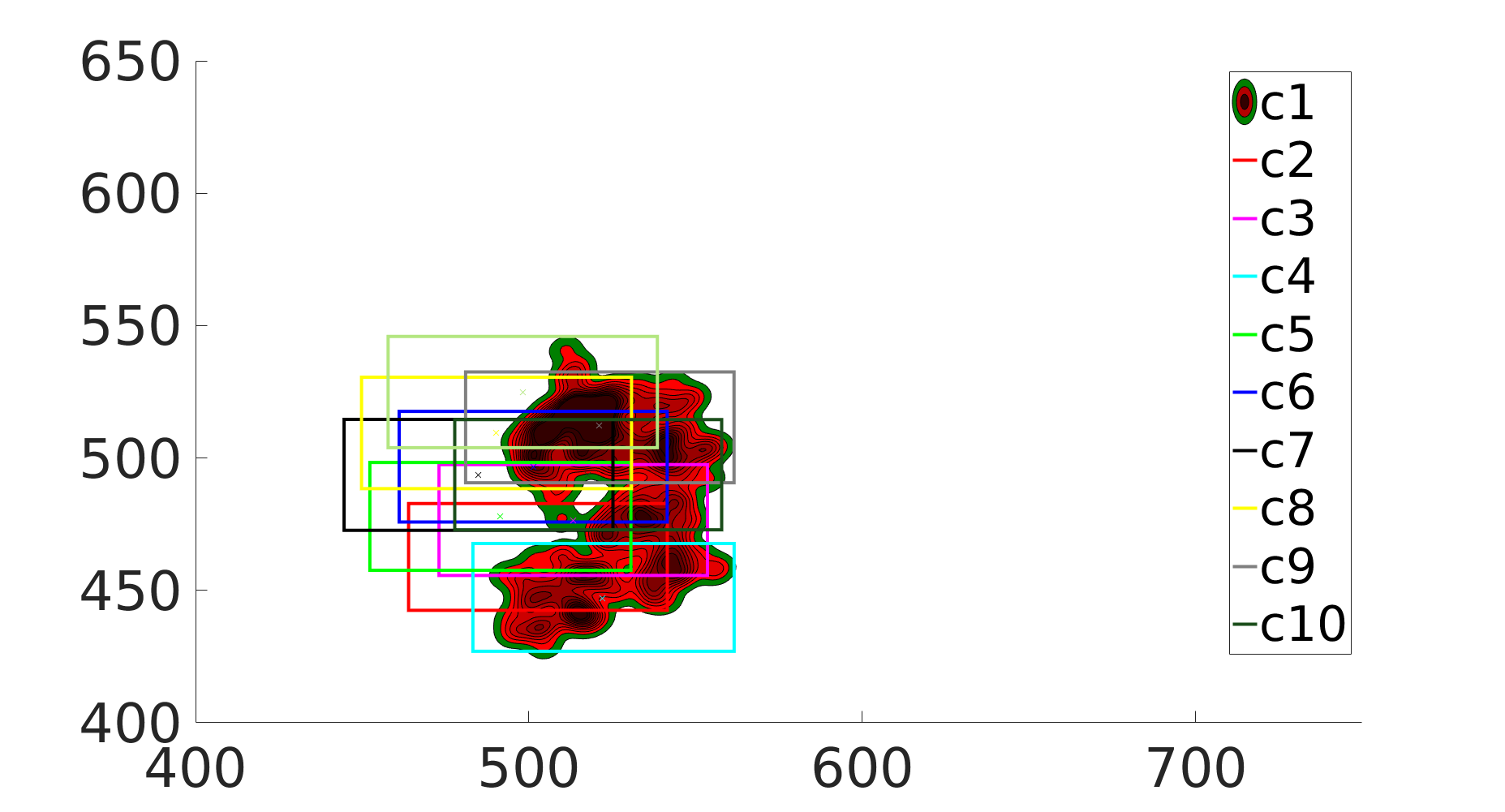}}
	\subfloat[t = 3000]{\includegraphics[width=0.25\textwidth]{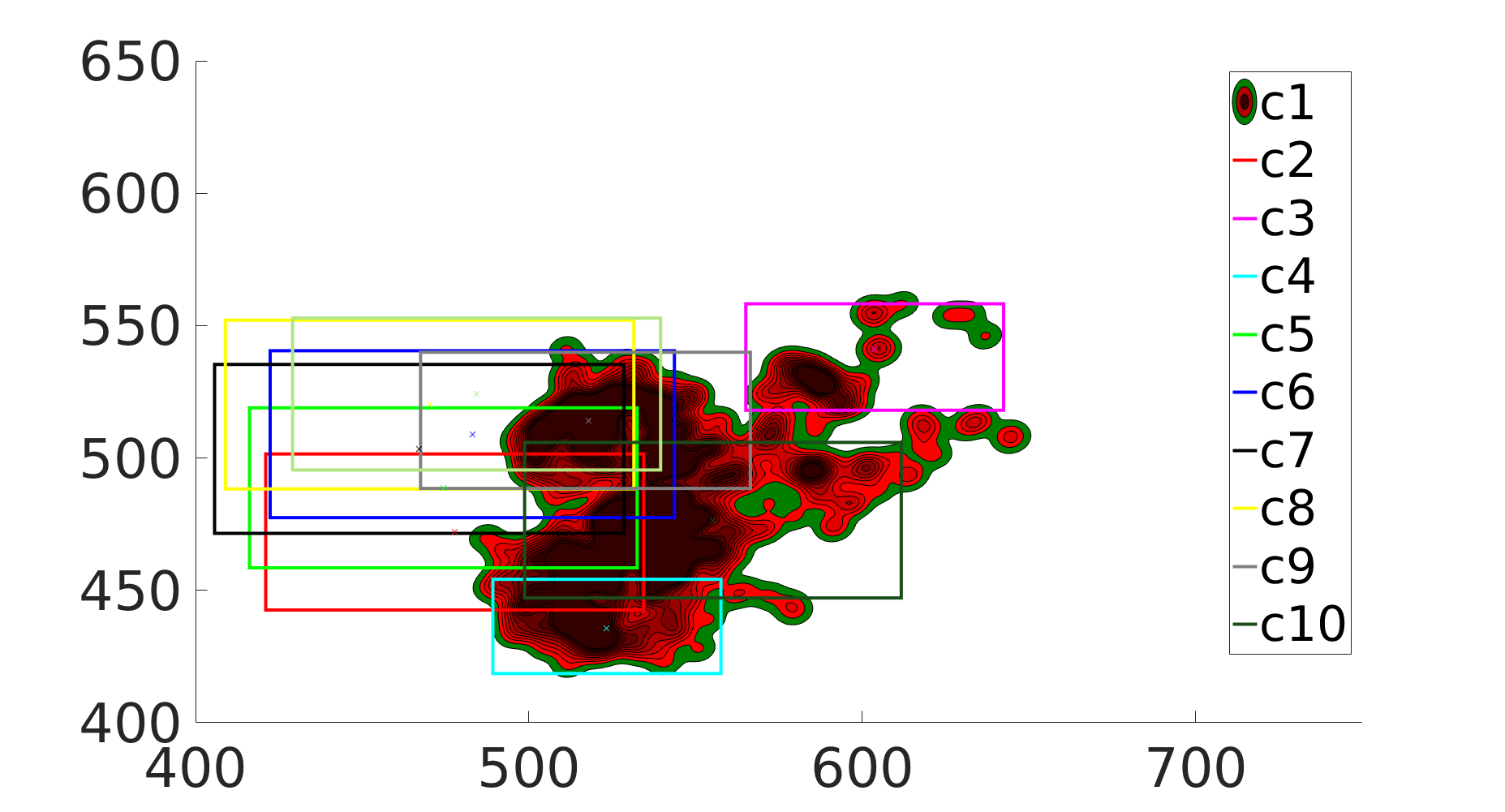}}
	\subfloat[t=  4000]{\includegraphics[width=0.25\textwidth]{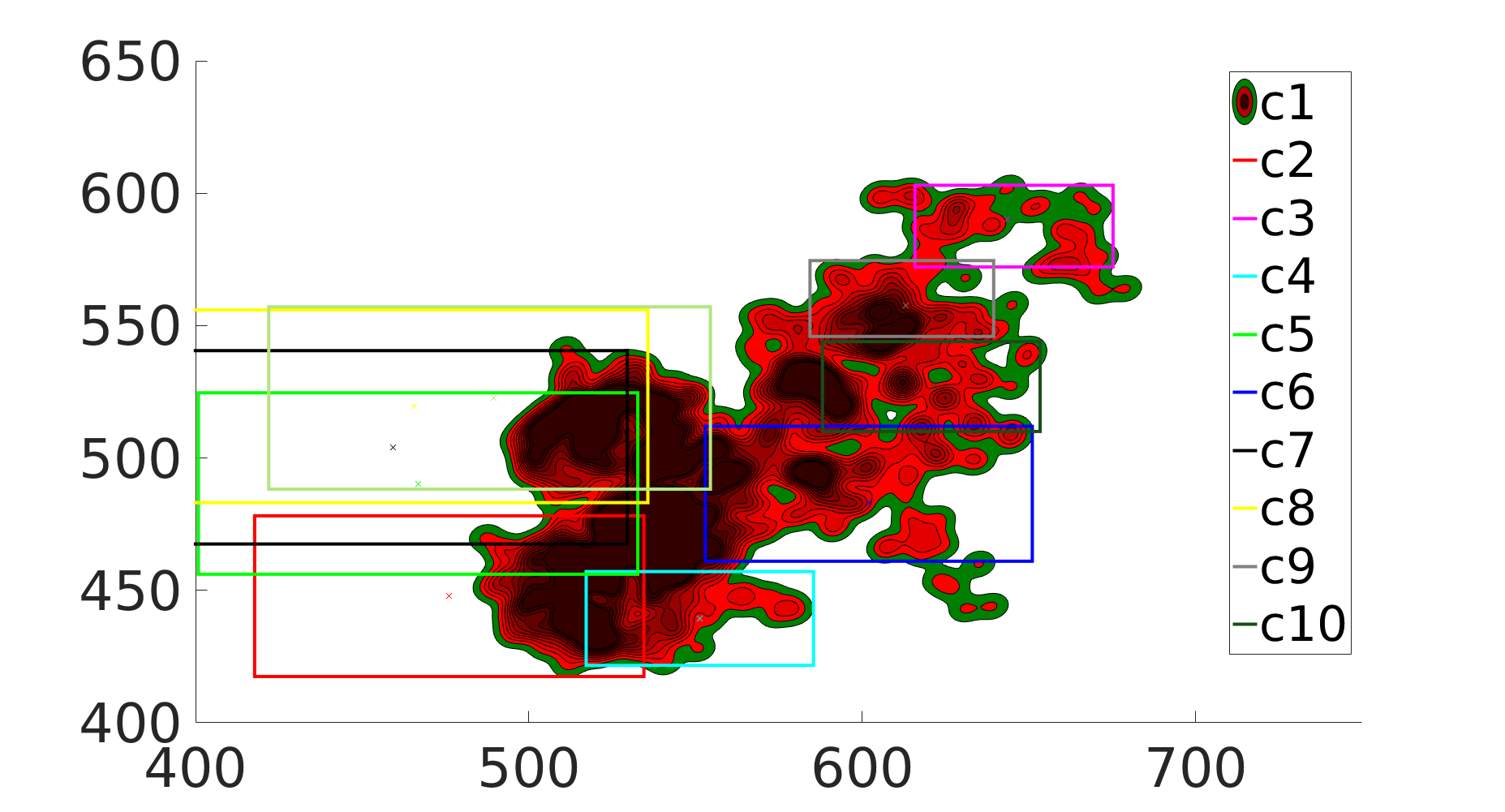}}
	\subfloat[t = 6000]{\includegraphics[width=0.25\textwidth]{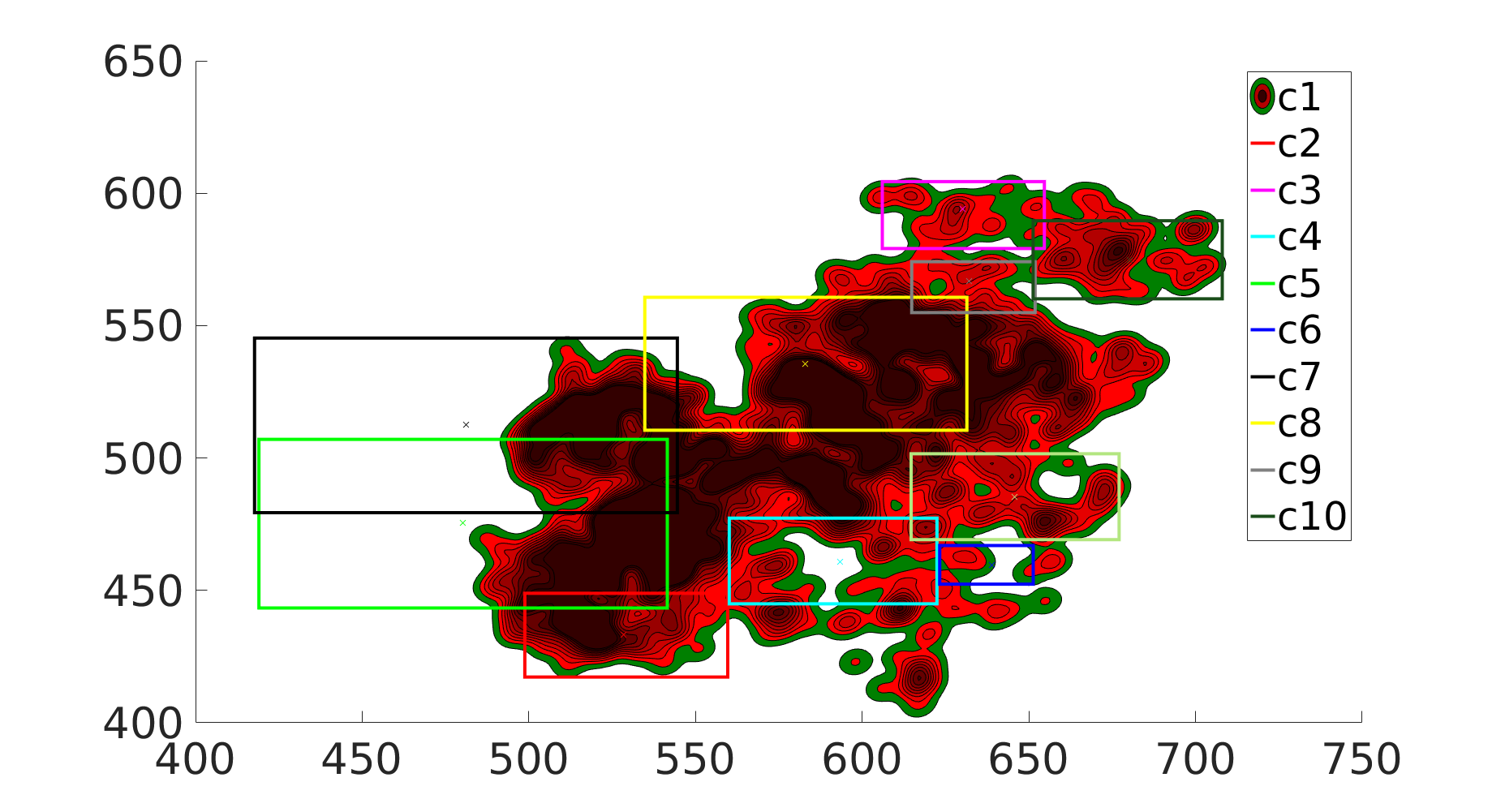}}
  \caption{Simulation result shows the field of view of each UAV on the ground in a) t = 1000, b) t = 3000, c) t= 4000, and d) t = 6000.}
  \label{F.Simulationc}
\vspace{-15pt}
\end{figure*}

\begin{figure*}[t]
 \centering
	\subfloat[t = 1000]{\includegraphics[width=0.25\textwidth]{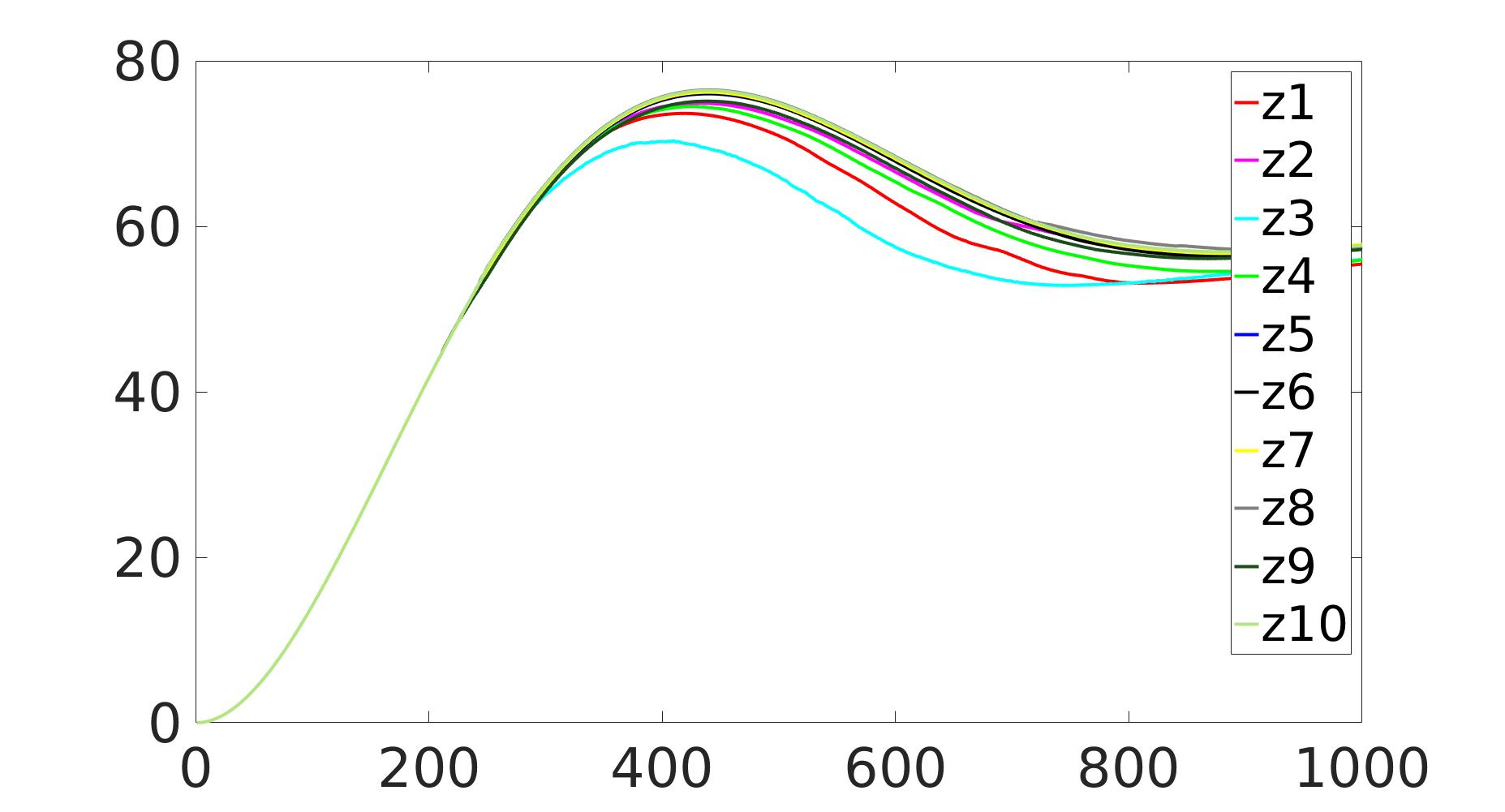}}
	\subfloat[t = 3000]{\includegraphics[width=0.25\textwidth]{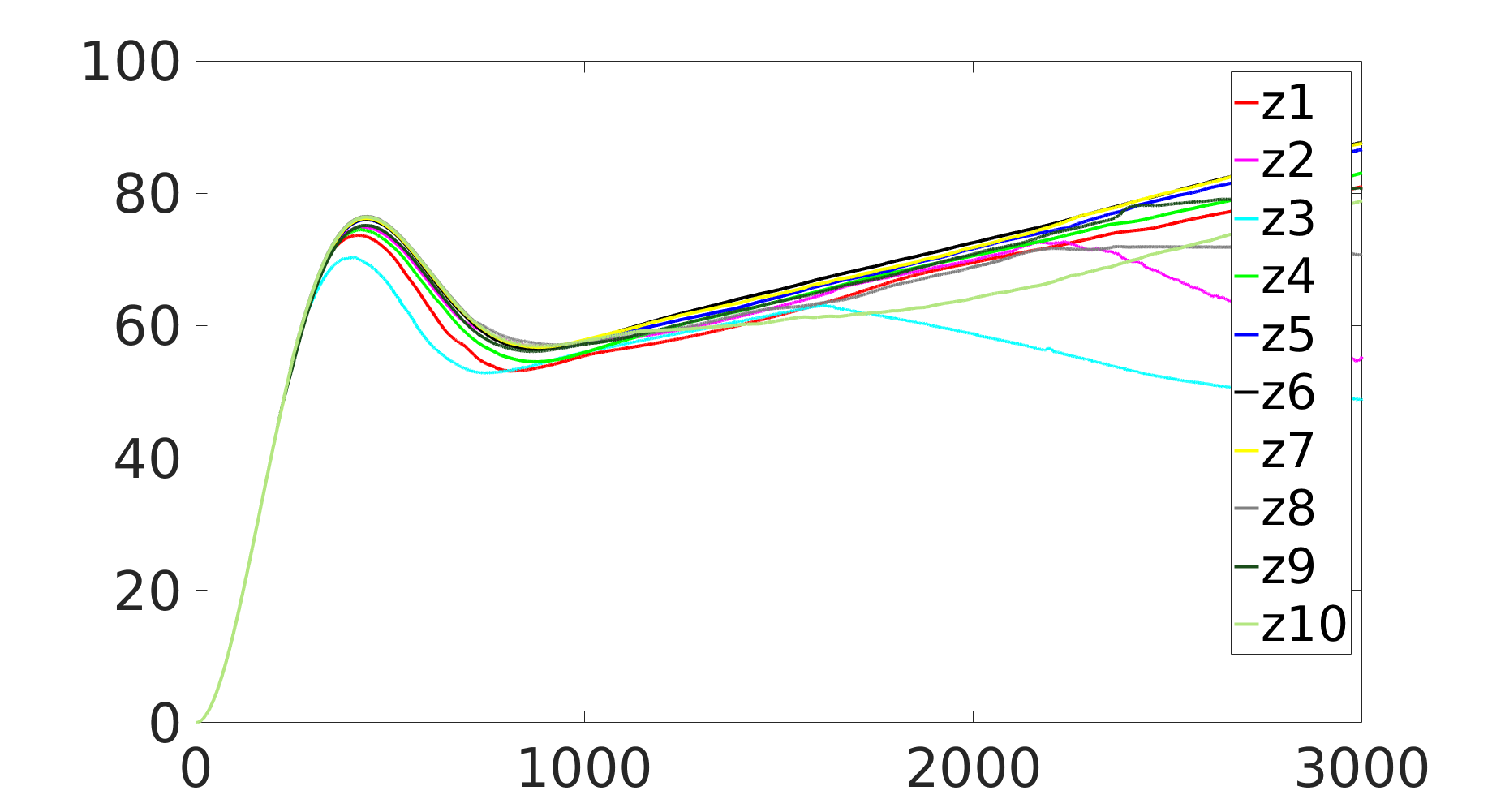}}
	\subfloat[t=  4000]{\includegraphics[width=0.25\textwidth]{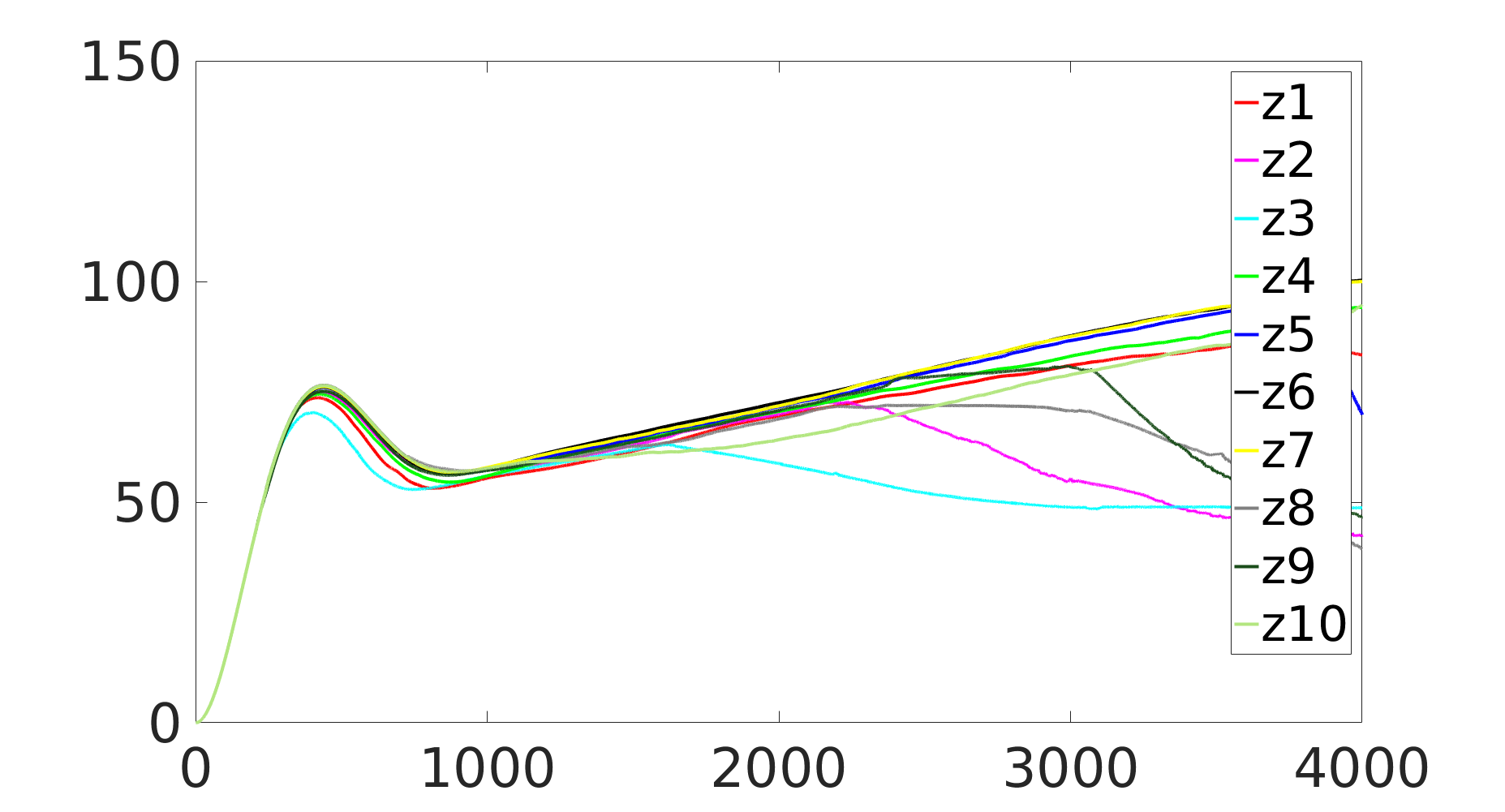}}
	\subfloat[t = 6000]{\includegraphics[width=0.25\textwidth]{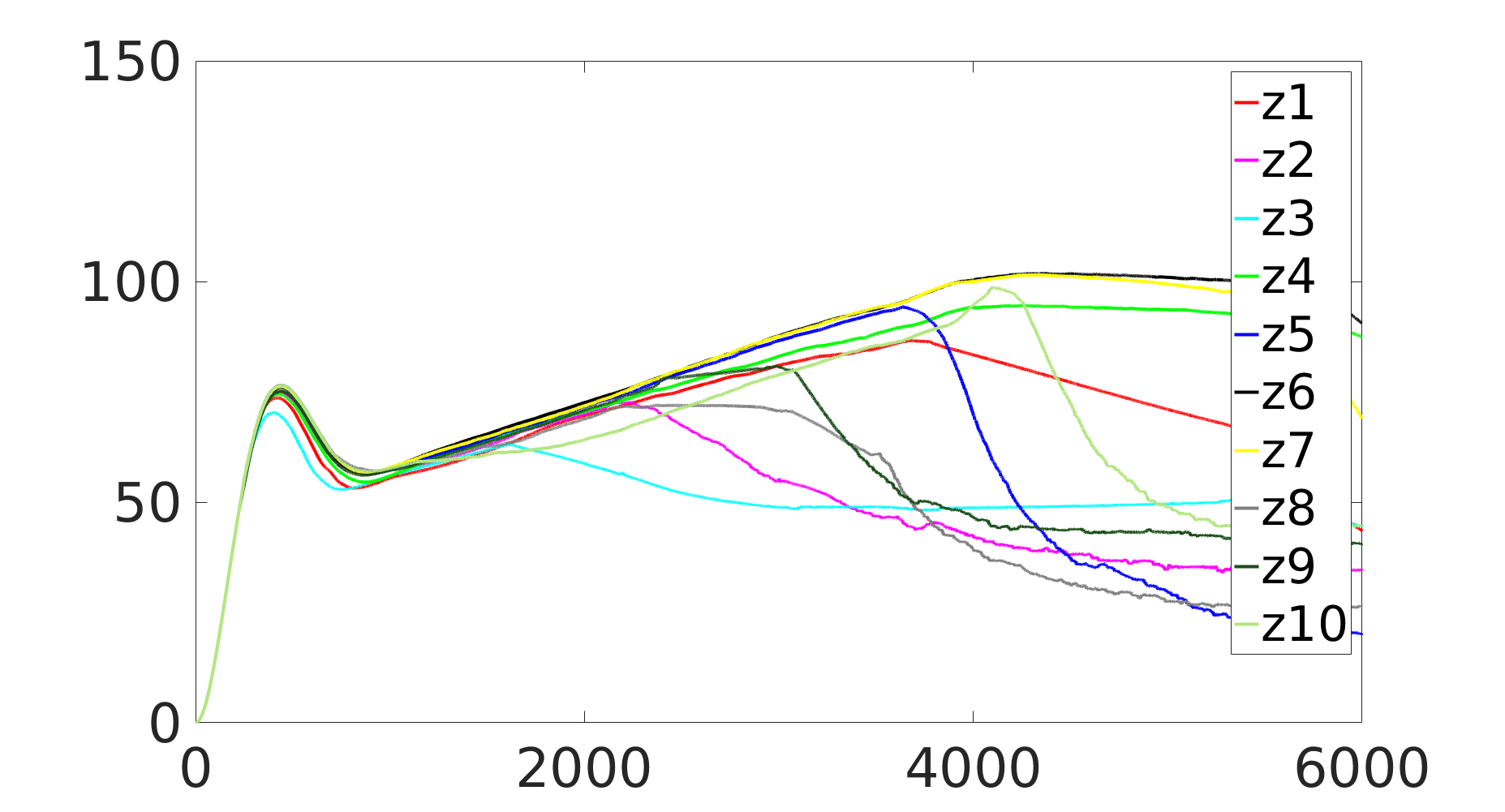}}
  \caption{Plot shows the altitude of each UAV on the ground in a) t = 1000, b) t = 3000, c) t= 4000, and d) t = 6000.}
  \label{F.Simulationz}
\vspace{-15pt}
\end{figure*}

\subsection{Coverage \& tracking control}
Based on the artificial potential field approach~\cite{schwager2011eyes,ge2000new, La_RAS_2012}, each UAV is distributedly controlled by a negative gradient (gradient descent) of the objective function $O$ in equation (\ref{3_coverobjective}) with respect to its pose $p_{i} = [c_{i}, z_{i}]^{T}$ as follows:
\begin{equation}\label{4.gradientcontrol}
\begin{aligned}
	\ u_{i} &= -k_{s}\frac{\partial O}{\partial p_{i}}, \\
\end{aligned}
\end{equation}
where $k_{s}$ is the proportional gain parameter. Then the lateral position and altitude of each UAV is controlled by taking the partial derivatives of the objective function $O$ as follows:
\begin{equation}\label{4.gradientcomponent}
\begin{aligned}
	\frac{\partial O}{\partial c_{i}} &= \sum_{k=1}^{4}\int \limits_{Q(t) \cap l_{ki}}(h_{N_{q}} - h_{N_{q}\setminus i})n_{ki}\kappa\Delta Idq, \\
	\frac{\partial O}{\partial z_{i}} &= \sum_{k=1}^{4}\int \limits_{Q(t) \cap l_{ki}}(h_{N_{q}} - h_{N_{q}\setminus i})tan \theta_{T}n_{ki}\kappa\Delta Idq,	\\
	&- \int \limits_{Q(t) \cap B_{i}} \frac{2h_{N_{q}}^{2}}{ \frac{S_{1}}{b^{2}}(b-z_{i})^{3}}\kappa\Delta Idq,   \\
\end{aligned}
\end{equation}
where $h_{N_{q}} = (\sum_{i\in N_{q}}f(p_{i},q)^{-1}+w^{-1})^{-1}$, $N_{q}\setminus{i}$ denotes the coverage neighbor set excludes the UAV $i$. In (\ref{4.gradientcomponent}), the component in the first row allows the UAV to move along $x$-axis and $y$-axis of the wildfire area which has $\Delta I$ is larger, while reduce the coverage intersections with other UAVs. The components in the second row allows the UAV to change its altitude along the $z$-axis to trade off between cover larger FOV (the first component) over the wildfire and to have a better resolution of the fire fronts propagation (the second component).
From (\ref{4.gradientcomponent}), the desired virtual position $p_{d_{i}}$ will be updated to the potential field control component in lower level of the controller (see figure \ref{F.Controller}):
\begin{equation}\label{4.virtualposition}
\begin{aligned}
	\ p_{d_{i}}(k+1) &= p_{d_{i}}(k) -k\Delta u, \Delta u = (\frac{\partial O_{1}}{\partial c_{i}}, \frac{\partial O_{1}}{\partial z_{i}}).\\
\end{aligned}
\end{equation}

\begin{figure*}[h]
 \centering
  \includegraphics[width=\textwidth]{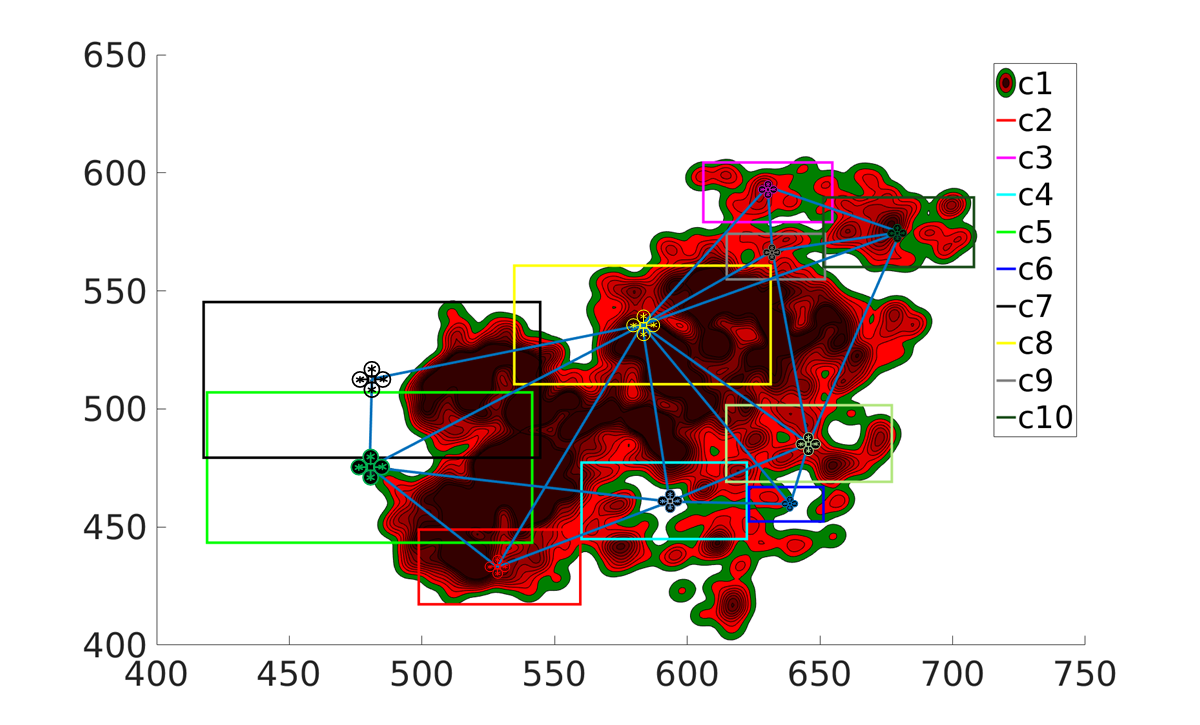}
  \caption{Result shows UAVs position and their FOV during the wildfire tracking.}
  \label{F.UAVFOV}
\vspace{0.5 cm}
\end{figure*}

\subsection{Potential field control}

Our approach is to use the artificial potential field to control each UAV to move to a desired position, and to avoid in-flight collision with other UAVs. We first create an attractive force to pull the UAVs to the initial rendezvous point $p_{r}$ by using a quadratic function of distance as the potential field, and take the gradient of it to yield the attractive force:

\begin{equation}\label{4.initialattractive}
\begin{aligned}
	\ U_{r}^{att} &= \frac{1}{2}k_{r}||p_{r} - p_{i}||^{2}\\
	\ u_{i}^{r} &= - \nabla U_{r}^{att} = -k_{r}(p_{i} - p_{r}). \\
\end{aligned}
\end{equation}

Similarly, the UAV move to desired virtual position, $p_{d_{i}}$, passed from equation (\ref{4.virtualposition}) in coverage \& tracking component, by using this attractive force:

\begin{equation}\label{4.virtualattractive}
\begin{aligned}
	\ U_{d}^{att} &= \frac{1}{2}k_{d}||p_{d_{i}} - p_{i}||^{2}\\
	\ u_{i}^{d} &= - \nabla U_{d}^{att} = -k_{d}(p_{i} - p_{d_{i}}). \\
\end{aligned}
\end{equation}

In order to avoid collision, we create repulsive forces from neighbors to push a UAV away if their distances become less than a designed safe distance $d$. Define the potential field for each neighbor UAV $j$ as:

\begin{equation}\label{4.repulsivePF}
\begin{aligned}
	\ U_{j}^{rep} &= 
	\begin{cases}
	\ \frac{1}{2}\nu(\frac{1}{||p_{j} - p_{i}||} - \frac{1}{d})^{2}, & if  \  ||p_{j} - p_{i}|| < d \\
    	 \ 0, & otherwise, \\
    	\end{cases}
\end{aligned}
\end{equation}
\noindent

where $\nu$ is a constant. The repulsive force can be attained by taking the gradient of the sum of the potential fields created by all neighbor UAVs as follows:

\begin{equation}\label{4.repulsive}
\begin{aligned}
	\ u_{i}^{rep} &= - \sum_{j \in N_{i}}a_{ij}\nabla U_{j}^{rep} \\
	&=\sum_{j \in N_{i}}\nu a_{ij}\Big(\frac{1}{||p_{j} - p_{i}||} - \frac{1}{d}\Big)\frac{1}{||p_{j}-p_{i}||^{3}}(p_{i} - p_{j})\\
	a_{ij} &=
	\begin{cases}
	\  1, &  if  \  ||p_{j} - p_{i}|| < d \\
    	 \ 0, &  otherwise. \\
    	\end{cases}
\end{aligned}
\end{equation}

From (\ref{4.initialattractive}), (\ref{4.virtualattractive}), and (\ref{4.repulsive}), we have the control equation for the lower control component:

\begin{equation}\label{4.lowercomponent}
\begin{aligned}
	\ u_{i} &= \sum_{j \in N_{i}}\nu a_{ij}\Big(\frac{1}{||p_{j} - p_{i}||} - \frac{1}{d}\Big)\frac{1}{||p_{j}-p_{i}||^{3}}(p_{i} - p_{j}) \\
	&- (1 - \zeta)k_{r}(p_{i} - p_{r}) -  \zeta k_{d}(p_{i} - p_{d_{i}}), \\
	\zeta_{i} &= 
	\begin{cases}
    	\ 1, &if  \    Q(t)\cap B_{i} \neq \varnothing \\
    	 \ 0, &if  \   otherwise. \\
    	\end{cases}
\end{aligned}
\end{equation}
\noindent

Note that, during the time the UAVs travel to the wildfire region, the coverage control component would not work because the sets $Q(t) \cap B_{i}$ and $Q(t) \cap l_{ki}$ are initially empty, so $\zeta_{i} = 0$. Upon reaching the waypoint region where the UAVs can sense the fire, $\zeta_{i} = 1$, that would cancel the potential force that draw the UAVs to the rendezvous point and let the UAVs track the fire fronts grow. 

\begin{figure*}[t]
 \centering
  \includegraphics[width=1.1\textwidth]{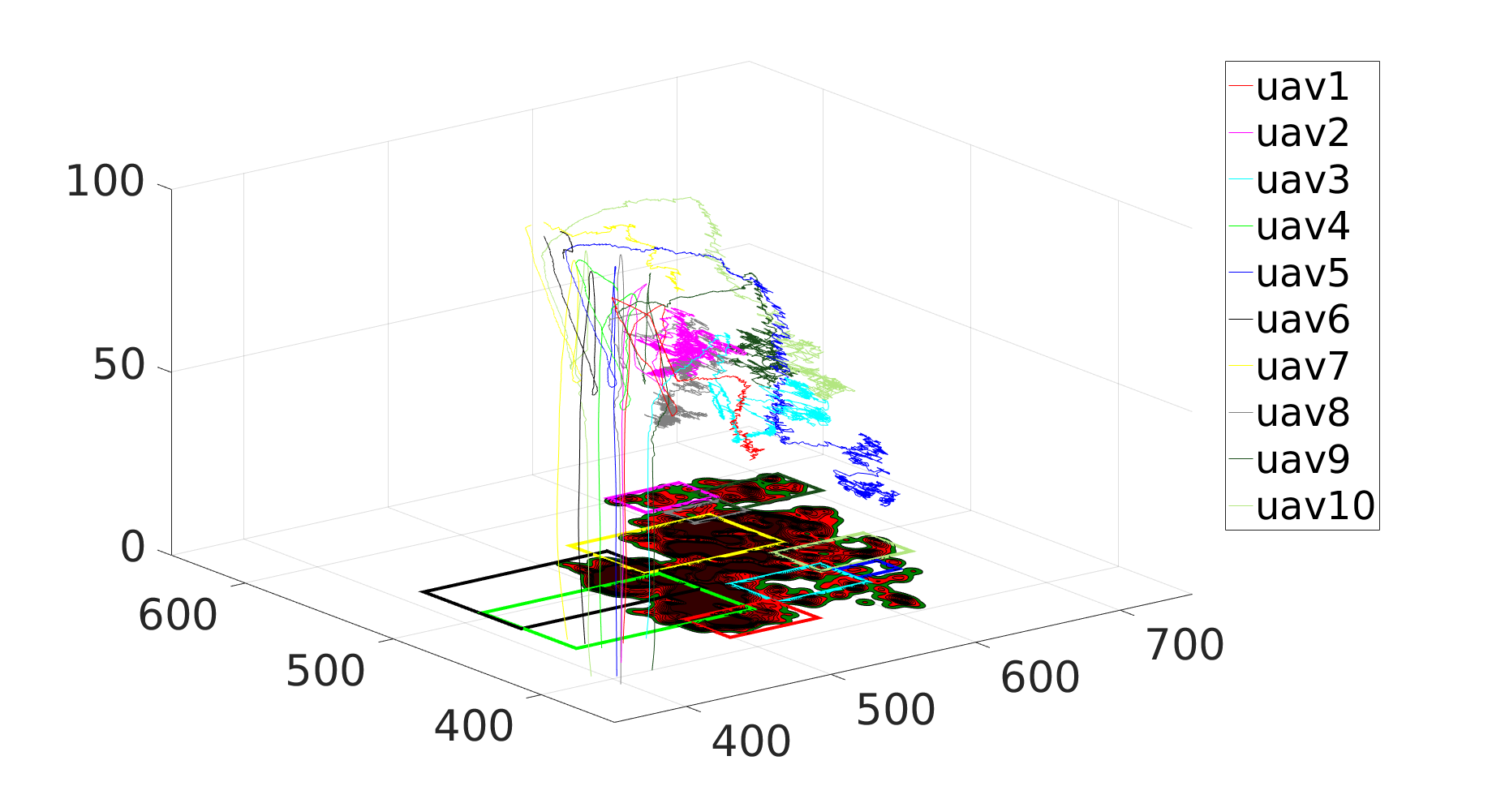}
  \caption{3D representation of the UAVs tracking the fire.}
  \label{F.Simulation3D}
\vspace{-15pt}
\end{figure*}

\section{Simulation}
Our simulation was conducted in a Matlab environment. We started with 10 UAVs on the ground ($z_{i} = 0$) from a fire fighting center with initial location arbitrarily generated around $[300, 300]^{T}$. The UAVs were equipped with identical cameras with focal length $b = 10$, area of one pixel $S_{1} = 10^{-4}$, half-angles $\theta_{1}=30 deg$, $\theta_{1}=45 deg$. The intensity sensitivity range of each camera was $[5, 100]^{T}$, and $\kappa = 10^{-3}$. The wildfire started with five initial fire front points near $[500, 500]^{T}$. The regulated mid-flame wind speed magnitude followed a Gaussian distribution with $\mu = 5 mph$ and $\sigma = 2$. The wind direction azimuth angle $\Theta$ also followed a Gaussian distribution with $\mu = \frac{\pi}{8}$ and $\sigma = 1$. The initial rendezvous point was $p_{r} = [500, 500, 60]^{T}$. The UAVs had a designed safe distance $d = 30$, and communication range $r = 100$.

We ran simulations in Matlab for 6000 time steps which yielded the result as shown in figure \ref{F.Simulationc} and \ref{F.Simulationz}. The UAVs came from the ground at $t = 0$ (Figure \ref{F.Simulationz}), and drove toward the wildfire region. Upon reaching the region near the initial rendezvous point at $[500, 500]^{T}$, the UAVs spread out to cover the entire wildfire (Figure \ref{F.Simulationc}-a). As the wildfire expanded, the UAVs fragment and follow the fire border regions (Figure \ref{F.Simulationc}-b, c, d). Note that the UAVs may not cover some regions with intensity $I = I_{max}$ (represented by black-shade color). Some UAVs may have low altitude if they cover region with small intensity $I$ (for example, UAV 5 in this simulation). The UAVs change altitude from $z_{i} \approx 60$ (Figure \ref{F.Simulationz}-a) to different altitudes (Figure \ref{F.Simulationz}-b, c, d), hence the area of the FOV of each UAV is different. It is obvious to notice that the UAVs attempted to follow the fire front propagation, hence satisfying the tracking objective. Figure \ref{F.UAVFOV} indicates the position of each UAV and its respective FOV in the last stage $t =  6000$. UAVs that are physical neighbors are connected with a blue line. We can see that most UAVs have sensing neighbors. Figure \ref{F.Simulation3D} shows the trajectory of each UAV in 3-dimensions while tracking the wildfire spreading north-east, and their current FOV on the ground.

\section{Conclusion}

In this paper, we presented a distributed control design for a team of UAVs that can collaboratively track a dynamic environment in the case of wildfire spreading. The UAVs can follow the border region of the wildfire as it keeps expanding, while still try to maintain coverage of the whole wildfire. The UAVs are also capable of avoiding collision, and flexible in deployment. The application could certainly go beyond the scope of wildfire tracking, as the system can work with any dynamic environment, for instance, oil spilling or water flooding. In the future, more work should be considered to research about the hardware implementation of the proposed controller. For example, we should pay attention to the communication between the UAVs under the condition of constantly changing topology of the networks, or the sensing endurance problem in hazardous environment. Also, we would like to investigate the relation between the speed of the UAVs and the spreading rate of the wildfire, and attempt to synchronize it. Multi-drone cooperative sensing \cite{La_SMCB_2013, La_SMCA_2015} and cooperative learning \cite{La_TCST_2015} for wildland fire mapping will be also considered.

\addtolength{\textheight}{-12cm}   





\section*{ACKNOWLEDGMENT}
This material is based upon work supported by the National Aeronautics and Space Administration (NASA) Grant No. NNX15AI02H issued through the Nevada NASA Research Infrastructure Development Seed Grant, and the National Science Foundation Cyber-Human Systems (NSF-CHS) Grant No. IIS - 152 8137.



\bibliographystyle{IEEEtran}
\bibliography{IROS2017_bibliography}

\begin{thebibliography}{10}
\providecommand{\url}[1]{#1}
\csname url@rmstyle\endcsname
\providecommand{\newblock}{\relax}
\providecommand{\bibinfo}[2]{#2}
\providecommand\BIBentrySTDinterwordspacing{\spaceskip=0pt\relax}
\providecommand\BIBentryALTinterwordstretchfactor{4}
\providecommand\BIBentryALTinterwordspacing{\spaceskip=\fontdimen2\font plus
\BIBentryALTinterwordstretchfactor\fontdimen3\font minus
  \fontdimen4\font\relax}
\providecommand\BIBforeignlanguage[2]{{%
\expandafter\ifx\csname l@#1\endcsname\relax
\typeout{** WARNING: IEEEtran.bst: No hyphenation pattern has been}%
\typeout{** loaded for the language `#1'. Using the pattern for}%
\typeout{** the default language instead.}%
\else
\language=\csname l@#1\endcsname
\fi
#2}}

\bibitem{nifc2017}
\BIBentryALTinterwordspacing
``National interagency fire center.'' [Online]. Available:
  \url{https://www.nifc.gov/fireInfo/fireInfo\_statistics.html}
\BIBentrySTDinterwordspacing

\bibitem{martinez2008computer}
J.~R. Martinez-de Dios, B.~C. Arrue, A.~Ollero, L.~Merino, and
  F.~G{\'o}mez-Rodr{\'\i}guez, ``Computer vision techniques for forest fire
  perception,'' \emph{Image and vision computing}, vol.~26, no.~4, pp.
  550--562, 2008.

\bibitem{stipanivcev2010advanced}
D.~Stipani{\v{c}}ev, M.~{\v{S}}tula, D.~Krstini{\'c}, L.~{\v{S}}eri{\'c},
  T.~Jakov{\v{c}}evi{\'c}, and M.~Bugari{\'c}, ``Advanced automatic wildfire
  surveillance and monitoring network,'' in \emph{6th International Conference
  on Forest Fire Research’, Coimbra, Portugal.(Ed. D. Viegas)}, 2010.

\bibitem{SujitP_2007}
P.~Sujit, D.~Kingston, and R.~Beard, ``Cooperative forest fire monitoring using
  multiple uavs,'' in \emph{Decision and Control, 2007 46th IEEE Conference
  on}, Dec 2007, pp. 4875--4880.

\bibitem{MerinoL_2006}
L.~Merino, F.~Caballero, J.~MartÃ­nez-de Dios, J.~Ferruz, and A.~Ollero, ``A
  cooperative perception system for multiple uavs: Application to automatic
  detection of forest fires,'' \emph{Journal of Field Robotics}, vol.~23, no.
  3-4, pp. 165--184, 2006.

\bibitem{cruz2016efficient}
H.~Cruz, M.~Eckert, J.~Meneses, and J.-F. Mart{\'\i}nez, ``Efficient forest
  fire detection index for application in unmanned aerial systems (uass),''
  \emph{Sensors}, vol.~16, no.~6, p. 893, 2016.

\bibitem{yuan2015survey}
C.~Yuan, Y.~Zhang, and Z.~Liu, ``A survey on technologies for automatic forest
  fire monitoring, detection, and fighting using unmanned aerial vehicles and
  remote sensing techniques,'' \emph{Canadian journal of forest research},
  vol.~45, no.~7, pp. 783--792, 2015.

\bibitem{YuanC_2015}
C.~Yuan, Z.~Liu, and Y.~Zhang, ``Uav-based forest fire detection and tracking
  using image processing techniques,'' in \emph{Unmanned Aircraft Systems
  (ICUAS), 2015 International Conference on}, June 2015, pp. 639--643.

\bibitem{merino2012unmanned}
L.~Merino, F.~Caballero, J.~R. Mart{\'\i}nez-de Dios, I.~Maza, and A.~Ollero,
  ``An unmanned aircraft system for automatic forest fire monitoring and
  measurement,'' \emph{Journal of Intelligent \& Robotic Systems}, vol.~65,
  no.~1, pp. 533--548, 2012.

\bibitem{MazaI_2011}
I.~Maza, F.~Caballero, J.~Capitan, J.~Martinez-de Dios, and A.~Ollero,
  ``Experimental results in multi-uav coordination for disaster management and
  civil security applications,'' \emph{Journal of Intelligent and Robotic
  Systems}, vol.~61, no. 1-4, pp. 563--585, 2011.

\bibitem{CasbeerD_2005}
D.~Casbeer, R.~Beard, T.~McLain, S.-M. Li, and R.~Mehra, ``Forest fire
  monitoring with multiple small uavs,'' in \emph{American Control Conference,
  2005. Proceedings of the 2005}, June 2005, pp. 3530--3535 vol. 5.

\bibitem{kumar2011cooperative}
M.~Kumar, K.~Cohen, and B.~HomChaudhuri, ``Cooperative control of multiple
  uninhabited aerial vehicles for monitoring and fighting wildfires,''
  \emph{Journal of Aerospace Computing, Information, and Communication},
  vol.~8, no.~1, pp. 1--16, 2011.

\bibitem{phan2008cooperative}
C.~Phan and H.~H. Liu, ``A cooperative uav/ugv platform for wildfire detection
  and fighting,'' in \emph{System Simulation and Scientific Computing, 2008.
  ICSC 2008. Asia Simulation Conference-7th International Conference on}.\hskip
  1em plus 0.5em minus 0.4em\relax IEEE, 2008, pp. 494--498.

\bibitem{glasa2011note}
J.~Glasa and L.~Halada, ``A note on mathematical modelling of elliptical fire
  propagation.'' \emph{Computing \& Informatics}, vol.~30, no.~6, 2011.

\bibitem{rothermel1972mathematical}
R.~C. Rothermel, ``A mathematical model for predicting fire spread in wildland
  fuels,'' 1972.

\bibitem{richards1990elliptical}
G.~D. Richards, ``An elliptical growth model of forest fire fronts and its
  numerical solution,'' \emph{International Journal for Numerical Methods in
  Engineering}, vol.~30, no.~6, pp. 1163--1179, 1990.

\bibitem{finney2004farsite}
M.~A. Finney \emph{et~al.}, \emph{FARSITE: Fire area simulator: model
  development and evaluation}.\hskip 1em plus 0.5em minus 0.4em\relax US
  Department of Agriculture, Forest Service, Rocky Mountain Research Station
  Ogden, UT, 2004.

\bibitem{williams2014modeling}
T.~M. Williams, B.~J. Williams, and B.~Song, ``Modeling a historic forest fire
  using gis and farsite,'' \emph{Mathematical \& Computational Forestry \&
  Natural Resource Sciences}, vol.~6, no.~2, 2014.

\bibitem{schwager2011eyes}
M.~Schwager, B.~J. Julian, M.~Angermann, and D.~Rus, ``Eyes in the sky:
  Decentralized control for the deployment of robotic camera networks,''
  \emph{Proceedings of the IEEE}, vol.~99, no.~9, pp. 1541--1561, 2011.

\bibitem{ge2000new}
S.~S. Ge and Y.~J. Cui, ``New potential functions for mobile robot path
  planning,'' \emph{IEEE Transactions on robotics and automation}, vol.~16,
  no.~5, pp. 615--620, 2000.

\bibitem{La_RAS_2012}
\BIBentryALTinterwordspacing
H.~M. La and W.~Sheng, ``Dynamic target tracking and observing in a mobile
  sensor network,'' \emph{Robotics and Autonomous Systems}, vol.~60, no.~7, pp.
  996 -- 1009, 2012. [Online]. Available:
  \url{http://www.sciencedirect.com/science/article/pii/S0921889012000565}
\BIBentrySTDinterwordspacing

\bibitem{La_SMCB_2013}
------, ``Distributed sensor fusion for scalar field mapping using mobile
  sensor networks,'' \emph{IEEE Transactions on Cybernetics}, vol.~43, no.~2,
  pp. 766--778, April 2013.

\bibitem{La_SMCA_2015}
H.~M. La, W.~Sheng, and J.~Chen, ``Cooperative and active sensing in mobile
  sensor networks for scalar field mapping,'' \emph{IEEE Transactions on
  Systems, Man, and Cybernetics: Systems}, vol.~45, no.~1, pp. 1--12, Jan 2015.

\bibitem{La_TCST_2015}
H.~M. La, R.~Lim, and W.~Sheng, ``Multirobot cooperative learning for predator
  avoidance,'' \emph{IEEE Transactions on Control Systems Technology}, vol.~23,
  no.~1, pp. 52--63, Jan 2015.

\end{thebibliography}

\end{document}